\documentclass[10pt,twocolumn,letterpaper]{article}

\usepackage{3dv}
\usepackage{times}
\usepackage{epsfig}
\usepackage{graphicx}
\usepackage{amsmath}
\usepackage{amssymb}
\usepackage{multirow}
\usepackage{authblk}

\threedvfinalcopy 

\def\threedvPaperID{267} 

\ifthreedvfinal\fi
\begin{document}

\title{SVMAC: Unsupervised 3D Human Pose Estimation from a Single Image with Single-view-multi-angle Consistency}

\newcommand*{\affaddr}[1]{#1}
\newcommand*{\affmark}[1][*]{\textsuperscript{#1}}
\newcommand*{\email}[1]{\texttt{#1}}
\author{
Yicheng Deng\affmark[1], Cheng Sun\affmark[2], Jiahui Zhu\affmark[1], Yongqi Sun\affmark[1*]\\
\affaddr{\affmark[1]Beijing Jiaotong University}\\
\affaddr{\affmark[2]Kyushu University}\\
\email{\tt\small yqsun@bjtu.edu.cn}
}

\maketitle
\thispagestyle{empty}
\pagestyle{empty}

\begin{abstract}
Recovering 3D human pose from 2D joints is still a challenging problem, especially without any 3D annotation, video information, or multi-view information. In this paper, we present an unsupervised GAN-based model consisting of multiple weight-sharing generators to estimate a 3D human pose from a single image without 3D annotations. In our model, we introduce single-view-multi-angle consistency (SVMAC) to significantly improve the estimation performance. With 2D joint locations as input, our model estimates a 3D pose and a camera simultaneously. During training, the estimated 3D pose is rotated by random angles and the estimated camera projects the rotated 3D poses back to 2D. The 2D reprojections will be fed into weight-sharing generators to estimate the corresponding 3D poses and cameras, which are then mixed to impose SVMAC constraints to self-supervise the training process. The experimental results show that our method outperforms the state-of-the-art unsupervised methods on Human 3.6M and MPI-INF-3DHP.
Moreover, qualitative results on MPII and LSP show that our method can generalize well to unknown data.
\end{abstract}

\begin{figure*}[htbp]
\centering
\includegraphics[width=0.9\textwidth]{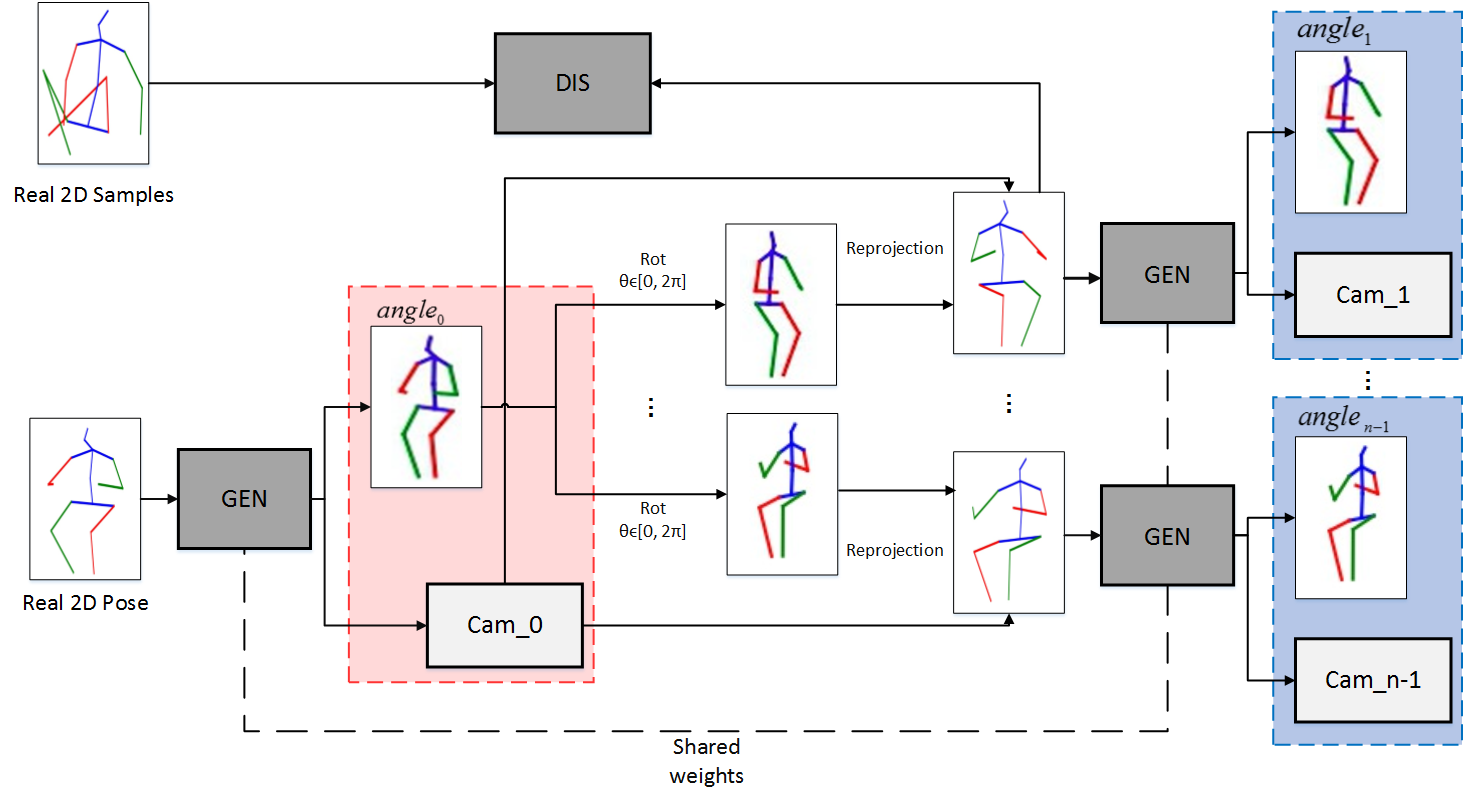}
\caption{The main structure of our adversarial training framework. The generators involved in our model share weights. 
The random reprojection of the estimated 3D pose are fed to the weight-sharing generator to be lifted to 3D again, allowing the network to impose SVMA consistency constraints. The 2D reprojection or a real 2D pose is fed to a discriminator for discrimination.} 
\label{netStruc}
\end{figure*}

\section{Introduction}
3D human pose estimation from monocular images has always been a problem in computer vision\cite{0Computational, 1983Model} with numerous applications such as motion recognition, virtual reality, and human-computer interaction. Although some currently presented fully supervised and weakly supervised methods based on deep learning have achieved good results\cite{2017Exploiting, 2017A, 2017VNect, ferrari_integral_2018,  2017Lifting}, these methods have two problems. Firstly, most of them learn a simple correspondence from 2D to 3D, which usually cannot be generalized to unknown actions and camera positions well. Secondly, these methods usually require much 3D annotation data, while there are currently few datasets with 3D annotations, especially for pose datasets in the wild, which is extremely difficult to be labeled with 3D annotation. 
Hence, the study of unsupervised methods for 3D human pose estimation is of great significance. 
Recently, several unsupervised methods have been proposed which don't require any 3D data. Some of them estimate 3D poses based on monocular images\cite{2020Unsupervised,2018Unsupervised,2020Kundu}, while the performance remains to be improved. Some of them use multi-view information and achieve good results\cite{2020Weakly,2019Self,2019WeaklyR,2020canonpose}, however setting up multi-view cameras in the wild is also extremely difficult.

In this paper, we propose an unsupervised adversarial training method for 3D human pose and camera estimation from 2D joint locations extracted from a single image. Figure \ref{netStruc} shows our training pipeline. 
In our model, a generator named $GEN$ is used to estimate a 3D pose and a camera simultaneously from an input 2D pose, and then we can reproject the estimated 3D pose to obtain the corresponding 2D pose.

Consider that a plausible 3D pose can be rotated by random angles and then be reprojected to obtain reasonable 2D poses. We propose single-view-multi-angle consistency(SVMAC) to improve the estimation accuracy. 
Specifically, we use generators that share weights with $GEN$ to impose SVMAC constraint. The estimated 3D pose obtained from $GEN$ is rotated from multiple angles and projected back to 2D reprojections. Then the 2D reprojections are fed to weight-sharing generators, which output the corresponding 3D poses and cameras. Since the estimated 3D poses from different angles are from a single view, we define the SVMAC loss, which will be described in detail in Section \ref{svmasec}. 

For the discriminator of our model, the input is the 2D reprojections or the 2D poses sampled from the real distribution. The discriminator aims to determine whether the 2D reprojections are from the real pose distribution, making our model learn a mapping of distribution from 2D poses to 3D poses, instead of a simple 2D-3D correspondence.

To verify the effectiveness of our method, we perform experiments on four datasets Human3.6M\cite{2014Human3}, MPI-INF-3DHP\cite{2017Monocular}, MPII\cite{2014Human}, and Leed Sports Pose(LSP)\cite{2010Clustered}. Results show that our method outperforms state-of-the-art methods, and the ablation studies on Human 3.6M and MPI-INF-3DHP datasets show that our SVMAC constraint significantly improves the performance of our model. In addition, the model trained on a specific dataset can also perform well on other datasets, which shows that our method has the excellent ability of generalization.

In summary, our contributions are as follows:
\begin{itemize}
\item We present the first unsupervised adversarial training method to simultaneously estimate a 3D pose and a camera from a 2D pose without requiring any other information.
\item Our method use weight-sharing networks to generate poses and cameras from multiple angles and mix them to impose single-view-multi-angle consistency (SVMAC) constraints, which significantly improves the estimation accuracy.
\item The experimental results show that our method outperforms the state-of-the-art methods, and can be generalized to unknown 3D human poses and cameras well.
\end{itemize}

\begin{figure*}[htbp]
\centering
\includegraphics[width=0.9\textwidth]{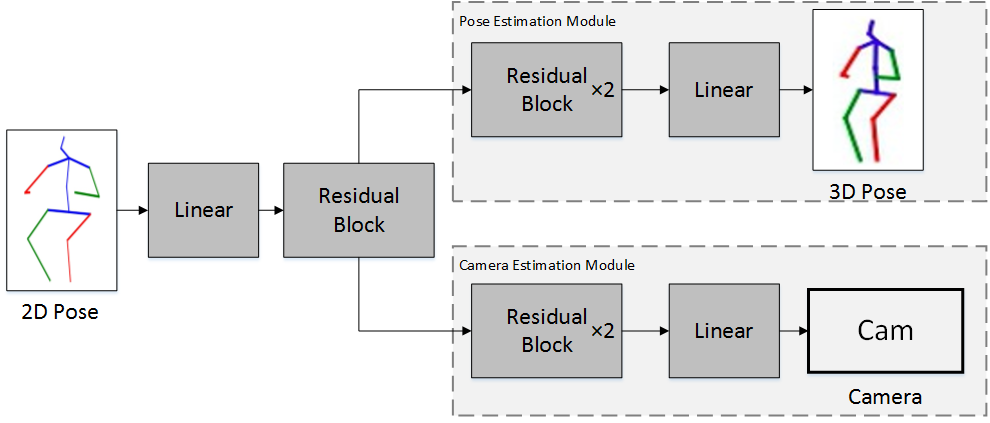}
\caption{Network structure of the generator. A fully connected layer and a shared residual block are used to upscale the input and extract its features. Then the network splits into two paths that predict the 3D pose and the camera, respectively. The upper path and lower path both have two residual blocks followed by a fully connected layer which outputs the 3D pose and the camera.}
\label{generator}
\end{figure*}

\section{Related work}
\noindent
{\bf Fully Supervised Methods} There are several methods that make full use of both 2D and 3D ground truth based on large datasets which contain millions of images with corresponding 3D pose annotations. Madadi et al.\cite{MADADI2020107472} use CNN-based 3D joint predictions to regress SMPL pose and shape parameters and then get the estimated 3D pose with these parameters. Sun et al.\cite{ferrari_integral_2018} propose an end-to-end model to regress a 3D human pose from 2D heat maps. Dushyant et al.\cite{2017VNect} present a CNN-based model, which regresses 2D and 3D joint coordinates and motion skeletons to produce a real-time stable 3D reconstruction of motion. In addition to the above end-to-end methods, there are also some methods whose estimation process includes two stages. The first stage is to perform 2D pose detection on a single image and predict its 2D joint coordinates\cite{2016Human,2017Realtime,carreira_human_2016,2016Stacked,2016DeepCut,2019Deep,toshev_deeppose_2014}, while the second stage is to predict 3D joint coordinates from the 2D joint coordinates through regression analysis or model fitting\cite{KIM2020107462,2018Unsupervised,2017A,2019RepNet}. Recently, some methods have been proposed for the second stage, and these methods aim to learn the 2D-3D correspondence with the given paired 2D and 3D data. Martinez et al.\cite{2017A} propose a simple but effective regression network to estimate a 3D human pose directly from a 2D pose, considered to be the baseline due to its simplicity and high accuracy estimation. Hossain et al.\cite{2017Exploiting} extend the baseline by employing a recurrent neural network for a human pose sequence. 
Although these methods have achieved outstanding performance, they require a lot of data with 3D annotations, and only work well on similar datasets.

\noindent
{\bf Weakly Supervised Methods} Weakly supervised methods only require limited 3D annotations or unpaired 2D-3D data. Zhou et al.\cite{2017Towards} propose a two-stage transfer learning method to generate 2D heat maps and regress the joint depths. Yang et al.\cite{yang_3d_2018} present an adversarial training method based on multiple representations. They introduce a discriminator that makes full use of RGB images, geometric representations, and heat maps. Drover et al.\cite{2018can} learn a mapping of distribution from 2D to 3D with the help of 2D projections based on a GAN\cite{goodfellow_i_generative_2014}. However, they require extra data generated by utilizing ground-truth 3D data for training. Considering the reprojection constraint, Wandt et al.\cite{2019RepNet} propose a GAN-based model named RepNet to estimate 3D pose and camera simultaneously, and use a discriminator to evaluate the estimated 3D pose and the corresponding KCS matrix. 
Although these methods somewhat solve the problem of generalization, they still require 3D annotations, which are time-consuming and labor-intensive to acquire.

\noindent
{\bf Unsupervised Methods} Unsupervised methods make full use of images or 2D data, and do not require any 3D annotation. Rhodin et al.\cite{2018Unsupervised1} propose an encoder-decoder model to perform 3D human pose estimation based on unsupervised geometry-aware representations. Their method requires multi-view 2D data to learn the appearance representation. Kudo et al.\cite{2018Unsupervised} consider that the random 2D reprojections are reasonable if the estimated 3D pose is accurate enough. Chen et al.\cite{2020Unsupervised} extend the method\cite{2018Unsupervised} by lifting the 2D reprojection to 3D again to impose geometric self-consistency constraints.  Kocabas et al.\cite{2019Self} adopt traditional methods to generate a 3D pose using multi-view 2D poses, and take it as self-supervised information. Kundu et al.\cite{2020Kundu} propose a differentiable and modular self-supervised method for 3D human pose estimation along with the discovery of 2D part segments from unlabeled video frames. 

However, these methods can't be generalized to unknown motions and camera positions well. In this paper, we propose an unsupervised adversarial training method to simultaneously estimate a 3D human pose and a camera from 2D joint locations extracted from a single image.

\section{Method}
In this section, we introduce our unsupervised method which lifts 2D joint locations to a 3D pose.
Let $x_{real}\in \mathbb{R}^{2N}=(x_{1},y_{1},x_{2},y_{2}...x_{N},y_{N})$ be 2D joint locations and $X_{real}\in \mathbb{R}^{3N}=(X_{1},Y_{1},Z_{1},X_{2},Y_{2},Z_{2},...,X_{N},Y_{N},Z_{N})$ be the real 3D pose, where $N$ represents the number of human joints. 
We take the hip joint as the root joint to align all 2D and 3D pose coordinates. Then, 
we assume a perspective camera, and we have
\begin{equation}
\begin{aligned}
x_{i} = f*X_{i}/Z_{i},\ \  i=1,2,...,N,
\label{xf}
\end{aligned}
\end{equation}
\begin{equation}
\begin{aligned}
y_{i} = f*Y_{i}/Z_{i},\ \  i=1,2,...,N,
\label{yf}
\end{aligned}
\end{equation}
where $f$ is the focal length of the camera. Here we assume $f=1$. 
And we assume that the distance from the camera to the 3D skeleton is $c$ = 10. 
Then we normalize 2D joint locations so that the average distance from other joints to the root joint (i.e., hip joint) is $1/c$, and correspondingly normalize the 3D poses so that the average distance from the other joints to the root joint is 1.

\begin{figure}[t]
\centering
\includegraphics[width=0.45\textwidth]{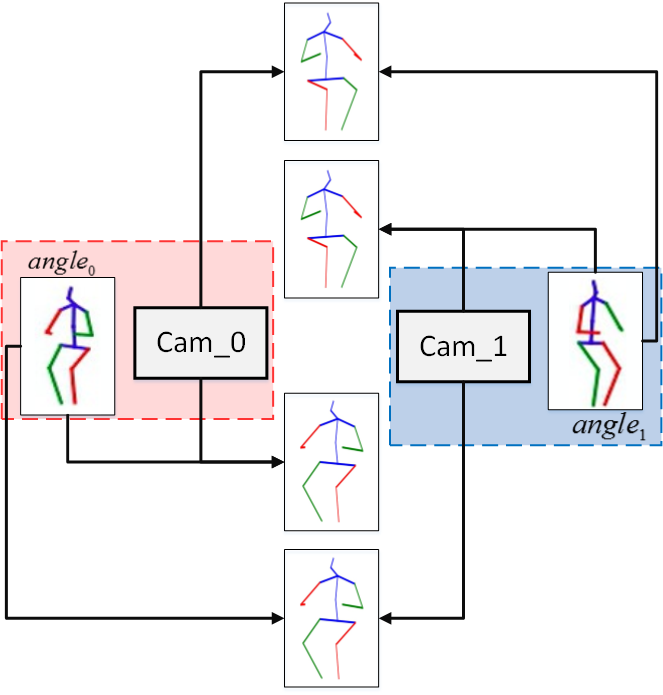}
\caption{An example of the single-view-multi-angle consistency in the case of 2 angles. The estimated 3D poses and cameras from different angles can be mixed to generate rich 2D reprojections, and the 2D reprojections from the same angle should be consistent.}
\label{SVMA}
\end{figure}

\subsection{3D pose and camera estimation}
Given the input $x_{real}$, we use a generator named $GEN$ to estimate the 3D pose and camera simultaneously.
\begin{equation}
\begin{aligned}
X_{pred},K=GEN(x_{real}).
\end{aligned}
\end{equation}
Specifically, $GEN$ contains two modules: pose estimation module and camera estimation module. 
Let $D=(d_{1},d_{2},...,d_{N})$ denotes the output of the pose estimation module, which represents the depth of each joint relative to the root joint. Then we have $Z = D+c$, and $X_{pred}=(X,Y,Z)$ by Eq.\ref{xf} and Eq.\ref{yf}.
Considering the difference among the average depths of the estimated 3D poses from different angles, the camera estimation module outputs a more simplified weak perspective camera $K\in \mathbb{R}^{2*3}$, which is used to mix the estimated 3D poses from the source angle and other angles to impose SVMAC constraints. 

Figure \ref{generator} shows the network structure of the generator. We first use a fully connected layer and a shared residual block to extract the feature of input 2D joint locations. Then, for the pose estimation module, we add two residual blocks after the shared residual block and finally add a fully connected layer to output a $N$-dimensional vector, while the camera estimation module contains two residual blocks after the shared residual block and a fully connected layer to output a $6$-dimensional vector. In addition, each layer is followed by batch-normalization\cite{2015Batch}, leaky ReLUs\cite{b_xu_n_wang_t_chen_m_li_empirical_2015}, and dropout\cite{2014Dropout}.

\subsection{Single-view-multi-angle consistency}
\label{svmasec}

Notice that a plausible 3D pose can be rotated by random angles and then be reprojected to obtain reasonable 2D poses. We consider a sequence of angles:
$angle_{0}$, $angle_{1}$, ..., $angle_{n-1}$,
where $n$ represents the number of angles. Let $\theta_{i}$ represent the difference between $angle_{i}$ and $angle_{0}$, $i=0,1,...,n-1,$ that is,
\begin{equation}
\begin{aligned}
\theta_{i} = 
angle_{i}-angle_{0},\ \ i=0,1,...,n-1,
\end{aligned}
\end{equation}
specifically $\theta_{0} = 0$.
Hence, $angle_{i}$
is obtained by rotating around the $y$-axis by $\theta_{i}$ radian from $angle_{0}$, which is the source angle of $X_{pred}$. Let $X_{rot}^{\theta_{i}}$ and $K^{\theta_{i}}$ represent the 3D pose and camera from the $angle_{i}$, we have


\begin{equation}
\begin{aligned}
X_{rot}^{\theta_{0}} = X_{pred},
\end{aligned}
\end{equation}
and
\begin{equation}
\begin{aligned}
K^{\theta_{0}} = K.
\end{aligned}
\end{equation}
Then we define the rotation matrix for $\theta_{i}$
\begin{equation}
\begin{aligned}
R^{\theta_{i}} = \left[\begin{array}
{ccc}
cos\theta_{i}&0&-sin\theta_{i} \\
0&1&0 \\
sin\theta_{i}&0&cos\theta_{i}
\end{array}\right], \ \ i=1,2,...,n-1.
\end{aligned}
\end{equation}
So the rotated 3D poses can be obtained by
\begin{equation}
\begin{aligned}
X_{rot}^{\theta_{i}}=(X_{pred}–[0,0,c])\ *\ R^{\theta_{i}} + [0,0,c], \\i=1,2,...,n-1.
\end{aligned}
\end{equation}
Through $X_{rot}^{\theta_{i}}$ and the estimated camera $K$, we have the rotated 2D reprojections
\begin{equation}
\begin{aligned}
x_{proj}^{\theta_{i}}=KX_{rot}^{\theta_{i}}, \ \ i=1,2,...,n-1.
\end{aligned}
\end{equation}
Then $x_{proj}^{\theta_{i}}$ is fed to a weight-sharing generator, we have
\begin{equation}
\begin{aligned}
\widetilde{X}_{rot}^{\theta_{i}},K^{\theta_{i}}=GEN(x_{proj}^{\theta_{i}}), \\ i=1,...,n-1.
\end{aligned}
\end{equation}
Since
$\widetilde{X}_{rot}^{\theta_{i}}$ and $X_{rot}^{\theta_{i}}$ should be similar, we define the loss function as
\begin{equation}
\begin{aligned}
\mathcal{L}_{3D}=\frac{1}{n-1}\sum_{i=1}^{n-1}\frac{1}{N}\left\|\widetilde{X}_{rot}^{\theta_{i}} - X_{rot}^{\theta_{i}}\right\|_{2}^{2}.
\label{3dloss}
\end{aligned}
\end{equation} 


Thus, we can define the single-view-multi-angle consistency(SVMAC), which means that the estimated poses and cameras from $n$ angles can be mixed to generate $n^2$ 2D reprojections, and the 2D reprojections from the same angle should be consistent. Figure \ref{SVMA} shows an example of the SVMA consistency in the case of two angles. The 2D reprojections can be calculated by 
\begin{equation}
\begin{aligned}
P_{a}^{\theta_{i}}=K^{\theta_{a}}\widetilde{X}_{rot}^{\theta_{i}}, \ \ a, i=0,1,...,n-1,
\end{aligned}
\end{equation} 
where
\begin{equation}
\begin{aligned}
\widetilde{X}_{rot}^{\theta_{0}} = X_{pred}. 
\end{aligned}
\end{equation} 
Then the loss function can be defined as
\begin{equation}
\begin{aligned}
\mathcal{L}_{svmac}=\frac{1}{n}\sum_{i=0}^{n-1}\sum_{a,b=1,a<b}^{m_{i}} \frac{1}{Nm_{i}} \left\|P_{a}^{\theta_{i}} - P_{b}^{\theta_{i}}\right\|_{2}^{2},
\label{svmaloss}
\end{aligned}
\end{equation} 
where $m_{i}$ is the number of 2D poses from $angle_{i}$. Note that the 2D poses involved in Eq. \ref{svmaloss} contain the input 2D pose at the source angle.



\subsection{Adversarial training}
To enhance the performance of our generators, we introduce a discriminator to determine the reality of the 2D reprojections to learn a mapping of the distribution from 2D poses to 3D poses instead of a simple 2D-3D correspondence. 
The input of the discriminator is the rotated 2D reprojection $x_{proj}^{\theta_{i}}$ or the 2D pose sampled from real samples $x_{sam}^{i}$. The output represents the probability that the current input is from real distribution.

The network structure of our discriminator includes a fully connected layer to increase the dimensionality of the input, two residual blocks and a fully connected layer to produce the output. For the activation functions we use leaky ReLUs.

\subsection{Loss functions}
For our GAN, the standard WGAN-GP loss function\cite{wgangp} is used, 
\begin{equation}
\begin{aligned}
\min \limits_{G}\max \limits_{D} \mathcal{L}_{adv}=\sum_{i=1}^{n-1}\mathbb{E}(D(x_{proj}^{\theta_{i}}))-\mathbb{E}(D(x_{sam}^{i}))\\+\lambda_{gp}(\nabla_{\hat{x}^{i}}(D(\hat{x}^{i})-1),
\end{aligned}
\end{equation} 
where $\hat{x}^{i}$ represents the linear combination of $x_{sam}^{i}$ and $x_{proj}^{\theta_{i}}$.

For the loss functions of camera, similar to \cite{2019RepNet}, each of the estimated cameras should satisfy
\begin{equation}
\begin{aligned}
K^{\theta_{i}}K^{\theta_{i}T}=s^2I_{2}, \ \ i=0,1,...,n-1,
\end{aligned}
\end{equation} 
where $s$ is the scale of the projection and $I_{2}$ is the 2*2 indentity matrix. Since $s$ equals to the largest singular value (or the $l_{2}$-norm) of $K^{\theta_{i}}$ and the trace of $K^{\theta_{i}}K^{\theta_{i}T}$ is the sum of the squared singular values, we can calculate $s$ for all angles as
\begin{equation}
\begin{aligned}
s=\sqrt{trace(K^{\theta_{i}}K^{\theta_{i}T})/2}, \ \ i=0,1,...,n-1.
\end{aligned}
\end{equation}
Then we can define the loss function of weak perspective camera as
\begin{equation}
\begin{aligned}
\mathcal{L}_{cam}=\sum_{i=0}^{n-1}\left\|\frac{2}{trace(K^{\theta_{i}}K^{\theta_{i}T})}K^{\theta_{i}}K^{\theta_{i}T}-I_{2}\right\|_{F},
\end{aligned}
\end{equation}
where$\left\|\cdot\right\|_{F}$ represents the Frobenius norm.


In addition, we consider the constraint of the bone length of the human body. There are several pairs of bones in the human body that have a symmetrical relationship in length. So we have the symmetric loss
\begin{equation}
\begin{aligned}
\mathcal{L}_{sym}=\frac{1}{q}\sum_{i}^q\left\|B_{i} - B_{i}'\right\|^2_{2},
\end{aligned}
\end{equation}
where $q$ is the number of pairs of bones that have a symmetrical relationship, $B_{i}$ and $B_{i}'$ are the $i$-th pair of two bones with a symmetrical relationship.

Then, we introduce another loss $\mathcal{L}_{angle}$ following the idea of Kudo et al.\cite{2018Unsupervised}. It guarantees that the $z$-components of the generated 3D pose will not be inverted. Similarly, we define the face orientation vector
$v=[v_x,v_y,v_z ]=j_{nose}-j_{neck}\in \mathbb{R}^3$ and shoulder orientation vector $w=[w_x,w_y,w_z]=j_{ls}-j_{rs}\in \mathbb{R}^3$, where $j_{nose},j_{neck},j_{ls},j_{rs}\in \mathbb{R}^3$ represent the 3D coordinates of the nose, neck, left shoulder, and right shoulder joints, respectively. According to the above mentioned constraints, the angle $\beta$ between $v$ and $w$ on the $z-x$ plane should satisfy
\begin{equation}
\begin{aligned}
\sin\beta =\frac{v_z w_x-v_x w_z}{\| v\|\| w\|}\geq 0.
\end{aligned}
\end{equation}
Thus, the loss function can be defined as
\begin{equation}
\begin{aligned}
\mathcal{L}_{angle}=max(0,-\sin\beta)=max(0,\frac{v_x w_z-v_z w_x}{\| v\|\| w\|} ).
\end{aligned}
\end{equation}

In the end, we define the total loss function of the generator as follows:
\begin{equation}
\begin{aligned}
\mathcal{L}=\mathcal{L}_{adv}+\lambda_{1}\mathcal{L}_{angle}+\lambda_{2}\mathcal{L}_{cam}+\\\lambda_{3}\mathcal{L}_{sym}+\lambda_{4}\mathcal{L}_{3D}+\lambda_{5}\mathcal{L}_{svmac},
\label{totalloss}
\end{aligned}
\end{equation}
where $\lambda_{i}$ for $1 \leq i \leq 5$
represent the weight coefficients of the loss terms, $\mathcal{L}_{3D}$ and $\mathcal{L}_{svmac}$ are calculated by Eq.\ref{3dloss} and Eq.\ref{svmaloss}, respectively.

\subsection{Training details}
As mentioned above, we use the standard WGAN-GP loss function and other loss functions to train our GAN. We use Adam optimizer\cite{2014Adam} for networks with learning rate of 5.5e-5, $beta_{1} = 0.7$ and $beta_{2}=0.9$. The loss weights are set as $\lambda_{1}$=1, $\lambda_{2}$=1, $\lambda_{3}$=0.01, $\lambda_{4}$=0.1 and $\lambda_{5}$=10 in Eq.\ref{totalloss}.


\section{Experiments and results}
In this section, we conduct several experiments to evaluate our model on the Human3.6M\cite{2014Human3}, MPI-INF-3DHP\cite{2017Monocular} datasets and show quantitative results. In addition, we also show qualitative results on the in-the-wild datasets MPII\cite{2014Human} and Leeds Sports Pose(LSP)\cite{2010Clustered}, where 3D ground truth data is not available.

\begin{table}[t]\small
\setlength\tabcolsep{8pt}
  \centering
  \caption{The results of 3D human pose estimation on the Human 3.6M dataset. GT and IMG denote the results obtained using ground-truth 2D joint locations and estimated 2D joint locations by SH/CPM\cite{2016Stacked, 2016CPM}, respectively.}
    \begin{tabular}{llll}
    \hline
    Type & Methods & GT & IMG \\
    \hline
    Full&Chen and Ramanan\cite{2016match} &57.5&82.7\\
    &Martinez et al.\cite{2017A}&37.1&52.1\\
    \hline
    Weak&3DInterpreter\cite{2016Single}&88.6&98.4\\
    &AIGN\cite{Tung_2017_ICCV}&79.0&97.2\\
    &RepNet\cite{2019RepNet}&38.2&65.1\\
    \hline
    Unsupervised&Kudo et al.\cite{2018Unsupervised}&73.2&110.2\\
    &Chen et al.\cite{2020Unsupervised}&58.0&-\\
    &Kundu et al.\cite{2020Kundu}&-&99.2\\
    \hline
    &Ours&\textbf{56.5}&\textbf{98.3}\\
    &Ours(SH/GT)&-&64.8\\
    \hline
    \end{tabular}%
  \label{h36m}%
\end{table}%

\subsection{Datasets and metrics}
\textbf{Human3.6M} Human3.6M is one of the largest 3D human pose datasets, consisting of 3.6 million 3D human poses. It contains video and MoCap data captured from 4 different viewpoints from 11 subjects performing typical activities such as directing, walking, sitting, etc. We evaluate the accuracy of pose estimation in terms of mean per joint position error(MPJPE) in millimeters after scaling and rigid alignment on the ground truth skeleton, i.e., P-MPJPE. We train our model on subjects S1, S5, S6, S7, S8 and evaluate it on subjects S9, S11.

\textbf{MPI-INF-3DHP} The MPI-INF-3DHP is a large human pose dataset consists of 3D data captures using a markerless multi-camera MoCap system. It contains both indoor and outdoor scenes and has eight actors performing several actions, which are more diverse than the Human 3.6M dataset. We evaluate valid images from the test-set containing 2929 frames following \cite{2018End} and report P-MPJPE, Percentage of Correct Keypoints (PCK) @150mm, and Area Under the Curve(AUC) computed for a range of PCK thresholds. For PCK and AUC, there are two cases of using or not using a rigid alignment.

\begin{figure*}[tbp]
\centering
\includegraphics[width=0.9\textwidth]{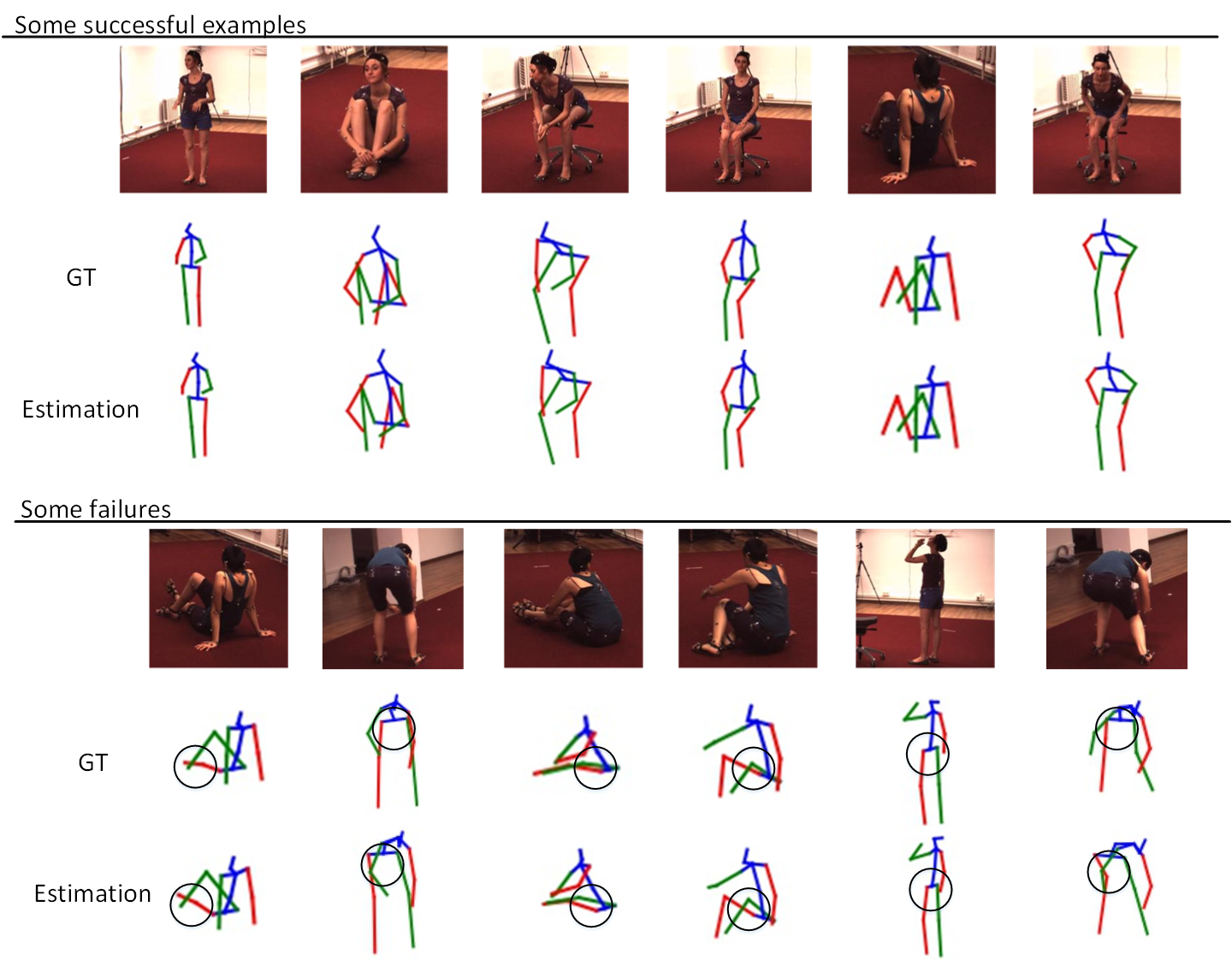}
\caption{Qualitative results on Human 3.6M dataset. The top shows some examples of successful reconstruction, and the bottom shows some failures. Each example includes image, ground-truth 3D pose and estimated 3D pose (top to bottom).}
\label{h36mqualr}
\end{figure*}

\subsection{Quantitative results on Human 3.6M}
Table \ref{h36m} shows the quantitative results on Human 3.6M. The lower value is better for P-MPJPE. The results show that our method outperforms state-of-the-art unsupervised methods by 2.6\%. In addition, we compare our method with several fully supervised and weakly supervised methods. It can be seen that our method even outperforms several weakly supervised methods and fully supervised methods. The results marked by SH/GT are obtained by training on 2D locations detected by stacked hourglass\cite{2016Stacked} and testing on 2D ground truth, and they indicate that our model works well for estimating the depth of human poses. Note that Chen et al.\cite{2020Unsupervised} achieve better performance by using temporal information and extra data for training, we will employ these information in novel ways to improve the performance in our future work.

\begin{table}[t]\footnotesize
\setlength\tabcolsep{4pt}
  \centering
  \caption{The results of 3D human pose estimation on the MPI-INF-3DHP dataset without using a scaling and rigid alignment. (\dag) denotes the method using temporal information, (\ddag) denotes the method using multi-view information, (+) denotes the method using extra data for training.}
    \begin{tabular}{lllcc}
    \hline
    Supervision&Methods & Training Data & \multicolumn{2}{c}{Absolute} \\
    &&&PCK&AUC \\
    \hline
    Full&VNect\cite{2017VNect}&H36M+MPI &76.6&40.4\\
    &Mehta\cite{2017Monocular}&MPI&72.5&36.9\\
    &Mehta\cite{2017Monocular}&H36M&64.7&31.7\\
    \hline
    Weak&SPIN\cite{2016Single}&Various&66.8&30.2\\
    &HMR\cite{2018End}&H36M+MPI&59.6&27.9\\

    \hline
    Unsupervised&Chen et al.\cite{2020Unsupervised}(\dag)(+)&H36M&64.3&31.6\\
    &Epipolar\cite{2019Self}(\ddag)&MPI&64.7&-\\
    \hline
    &Ours&H36M&\textbf{64.8}&31.6\\
    &Ours&MPI&\textbf{66.5}&\textbf{33.0}\\
    \hline
    \end{tabular}%
  \label{mpi3d1}%
\end{table}%

\begin{table}[t]\footnotesize
\setlength\tabcolsep{3pt}
  \centering
  \caption{The results of 3D human pose estimation on the MPI-INF-3DHP dataset using a scaling and rigid alignment, in which Various refers to the combination of datasets H36M, MPI-INF-3DHP and LSP.}
    \begin{tabular}{lllccc}
    \hline
    Supervision&Methods & Training Data &\multicolumn{3}{c}{Rigid Alignment} \\
    &&&PCK&AUC&P-MPJPE \\
    \hline
    Full&VNect\cite{2017VNect}&H36M+MPI &83.9&47.3&98.0\\
    &DenseRac\cite{2020DenseRaC}&H36M+MPI&86.3&47.8&89.8\\
    \hline
    Weak&SPIN\cite{2016Single}&Various&87.0&48.5&80.4\\
    &HMR\cite{2018End}&H36M+MPI&77.1&40.7&113.2\\
    \hline
    Unsupervised&PoseNet3D\cite{2020PoseNet3D}&H36M&81.9&43.2&102.4\\
    &Kundu et al.\cite{2020Kundu}&H36M&82.1&\textbf{56.3}&103.8\\
    &Kundu et al.\cite{2020Kundu}&MPI&84.6&\textbf{60.8}&93.9\\
    \hline
    &Ours&H36M&\textbf{86.9}&51.7&\textbf{80.3}\\
    &Ours&MPI&\textbf{86.6}&53.1&\textbf{79.8}\\
    \hline
    \end{tabular}%
  \label{mpi3d2}%
\end{table}%

\begin{table}[t]\small
\setlength\tabcolsep{6pt}
  \centering
  \caption{Ablation Studies.}
    \begin{tabular}{ccc}
    \hline
    &\multicolumn{2}{c}{P-MPJPE}\\
    Methods & Human 3.6M & MPI-INF-3DHP\\
    \hline
    w/o Disc&90.0&134.2\\
    \hline
    $n = 1$&70.9 & 98.7\\
    $n = 2$&56.8&79.8\\
    $n = 3$&56.5&78.5\\
    \hline
    $n = 2$ with & \multirow{2}*{\shortstack{35.4}} & \multirow{2}*{\shortstack{51.0}}\\
    5\% 3D Sup&&\\
    \hline
    \end{tabular}%
  \label{ablation}%
\end{table}%

\subsection{Quantitative results on MPI-INF-3DHP}

Table \ref{mpi3d1} and Table \ref{mpi3d2} show the qualitative results on MPI-INF-3DHP. Higher values of PCK and AUC signify better performance. Comparing with unsupervised methods which estimate a 3D pose from a single image, the experimental results show that our method outperforms state-of-the-art by 15.0\% in terms of P-MPJPE with training on MPI-INF-3DHP, and 21.6\% in terms of P-MPJPE with training on Human 3.6M, respectively. Besides, our method is even better than several unsupervised methods that use multi-view data or temporal data and extra data for training. In addition to comparing with the state-of-the-art unsupervised methods, we also show results from top fully supervised and weakly supervised methods. Hence, the experimental results show that our model can be applied to multiple datasets and achieves better performance for human pose estimation. 

As shown in Table \ref{mpi3d2}, we notice that there is only a minor difference of experimental results between the training datasets of MPI-INF-3DHP and Human3.6M. This indicates that our method can converge to a similar distribution of feasible human poses for both training sets, which also means our method can be well generalized to unseen data.

\subsection{Ablation studies}

In this section, we conduct ablation studies on Human 3.6M and MPI-INF-3DHP to evaluate the effectiveness of the adversarial training method and our single-view-multi-angle consistency in terms of P-MPJPE. The experimental results are shown in Table \ref{ablation}. 'w/o Disc' represents the experimental results when we train our model without using a discriminator. The results on Human 3.6M and MPI-INF-3DHP show that the adversarial training method helps learn a mapping from 2D poses to 3D poses. 

'$n=k$' represents the results when we use the SVMAC loss, see Eq.\ref{svmaloss}, in the case of $k$ angles.
Specially, the case of '$n = 1$' means that we train the model without the SVMAC loss.
For the case of $n=2$, 
The experimental results show that the SVMA consistency loss further improves the performance of our model by 20\% compared with $n=1$. Hence, we have the conclusion that the simple reprojection constraints and adversarial training are not enough for estimating reasonable 3D poses.
The SVMAC loss significantly improves the accuracy of our model for estimating plausible 3D poses and cameras.
In order to further explore the effectiveness of the value of $n$, we perform experiments in the case of $n=3$ as shown in Table \ref{ablation}. We can find that increasing the number of angles does not significantly improve performance, but makes training process much slower. Hence we set $n=2$ to train our model and compare it with other methods.

In addition, we also perform experiments on both datasets with 5\% ground truth 3D data for supervision, and the results are shown at the bottom of table 4. The experimental results show that our model can outperform most of weakly supervised methods and even fully supervised methods by using a little 3D ground truth for supervision.

\subsection{Qualitative results}
In this section, we conduct qualitative experiments on Human 3.6M dataset and Figure \ref{h36mqualr} shows several reconstruction examples. The results show that our method can perform very well for even more complicated actions. However, our model cannot work well for certain special views and scenes.

\begin{figure}[h]
\centering
\includegraphics[width=0.5\textwidth]{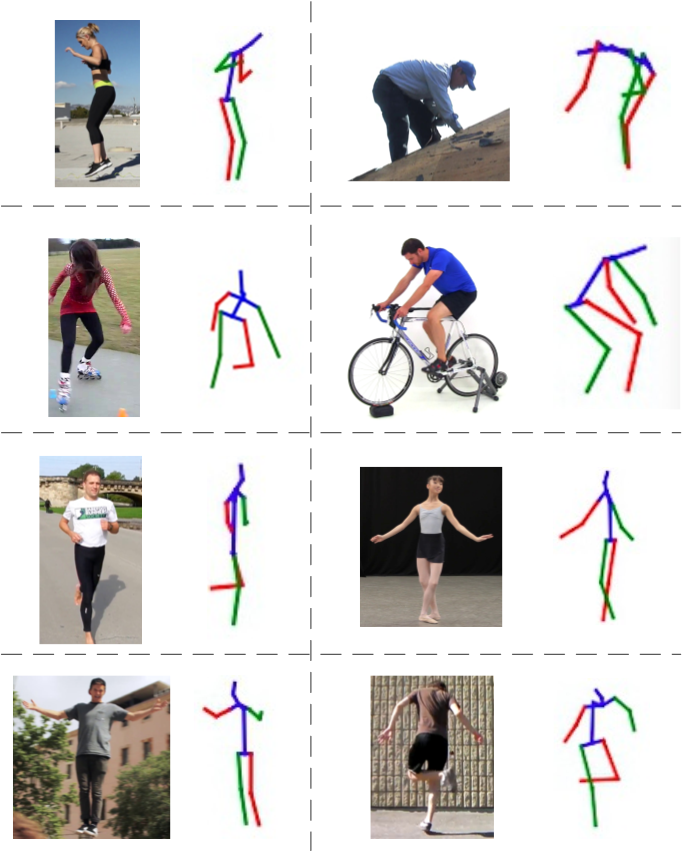}
\caption{Examples of reconstruction on the MPII dataset. Each example shows an image and the corresponding reconstructed 3D pose.}
\label{MPII}
\end{figure}

\begin{figure}[h]
\centering
\includegraphics[width=0.5\textwidth]{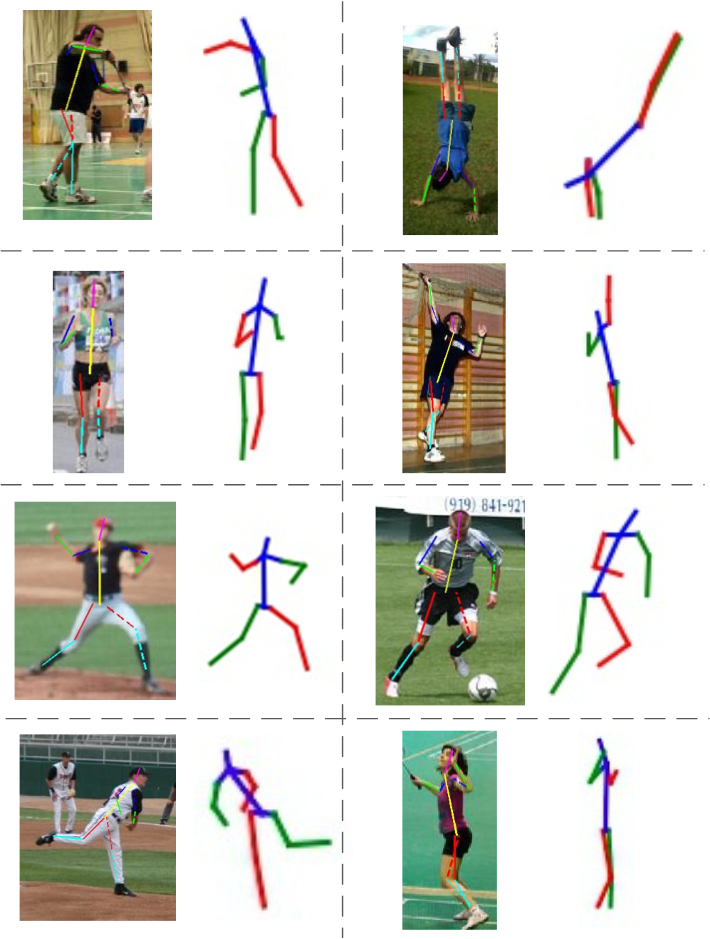}
\caption{Examples of reconstruction on the LSP dataset. Each example shows the image with overlaid 2D pose and the corresponding reconstructed 3D pose.}
\label{LSP}
\end{figure}

In order to verify the generalization of our model, we conduct experiments on MPII and LSP datasets without training on them but on Human 3.6M, and show several examples of estimated 3D poses in Figure \ref{MPII} and Figure \ref{LSP}, respectively. It can be seen that our model performs well with standard 2D pose datasets, which contain more complicated in-the-wild poses and actions that are not involved in the training dataset.

\section{Conclusion}
For 3D human pose estimation, the acquisition of 3D annotation data is time-consuming and expensive, which is still a difficult problem. 
In this paper, we present an unsupervised GAN-based model to estimate a 3D pose and a camera simultaneously from a 2D pose extracted from a single image without requiring any other information. 
Considering that a plausible 3D pose can be projected back to reasonable 2D poses even if it is rotated by random angles, we first propose single-view-multi-angle consistency(SVMAC).
We use weight-sharing generators to impose SVMAC constraints, which forces the estimated 3D pose reasonable from any angle. 
The experimental results show that our method outperforms the state-of-the-art unsupervised methods by 2.6\% on Human 3.6M and 15.0\% on MPI-INF-3DHP in terms of P-MPJPE. We also perform qualitative experiments on MPII and LSP, which demonstrate that our method can be generalized to unknown actions and camera positions well. In the future, we plan to improve the model's performance by applying it to video sequences.

{\small
\bibliographystyle{ieee}
\bibliography{main}   

}

\end{document}